\documentclass[journal,twoside,web]{utils/ieeecolor}

\usepackage{utils/jsen}
\usepackage{cite}
\usepackage{amsmath,amssymb,amsfonts}
\usepackage{algorithmic}
\usepackage{graphicx}
\usepackage{textcomp}
\usepackage{wrapfig}
\usepackage{balance}
\usepackage{soul}
\soulregister\cite7
\soulregister\ref7

\def\BibTeX{{\rm B\kern-.05em{\sc i\kern-.025em b}\kern-.08em
    T\kern-.1667em\lower.7ex\hbox{E}\kern-.125emX}}
\definecolor{abstractbg}{rgb}{0.89804,0.94510,0.83137}
\usepackage{color, colortbl}
\definecolor{lgreen}{rgb}{0.5,1,0.5}
\definecolor{lred}{rgb}{1,0.5,0.5}
\definecolor{lgray}{rgb}{0.88,0.88,0.88}
\definecolor{orangee}{rgb}{1.0,0.5,0.0}

\usepackage{siunitx}
\usepackage{hyperref}
\hypersetup{
    colorlinks=true,
    linkcolor=blue,
    filecolor=magenta,      
    urlcolor=cyan,
}

\begin{document}

\newpage

\title{
SmartDepthSync: Open Source Synchronized Video Recording System of Smartphone RGB and Depth Camera Range Image Frames with Sub-millisecond Precision}
\author{Marsel Faizullin, Anastasiia Kornilova, Azat Akhmetyanov, \\ Konstantin Pakulev, Andrey Sadkov and Gonzalo Ferrer\\
Skolkovo Institute of Science and Technology, Moscow, Russia
\thanks{
This research is based on the work supported by Samsung Research, Samsung Electronics.}
\thanks{The author is with Skolkovo Institute of Science and Technology.
 {\tt\small \{marsel.faizullin, anastasiia.kornilova, a.akhmetyanov, konstantin.pakulev, a.sadkov, g.ferrer\}@skoltech.ru}. }
\thanks{}
}

\IEEEtitleabstractindextext{%
\fcolorbox{abstractbg}{abstractbg}{%
\begin{minipage}{\textwidth}%
    \begin{wrapfigure}[13]{r}{2.5in}
    \vspace{-1.5mm}
    \includegraphics[width=2.4in]{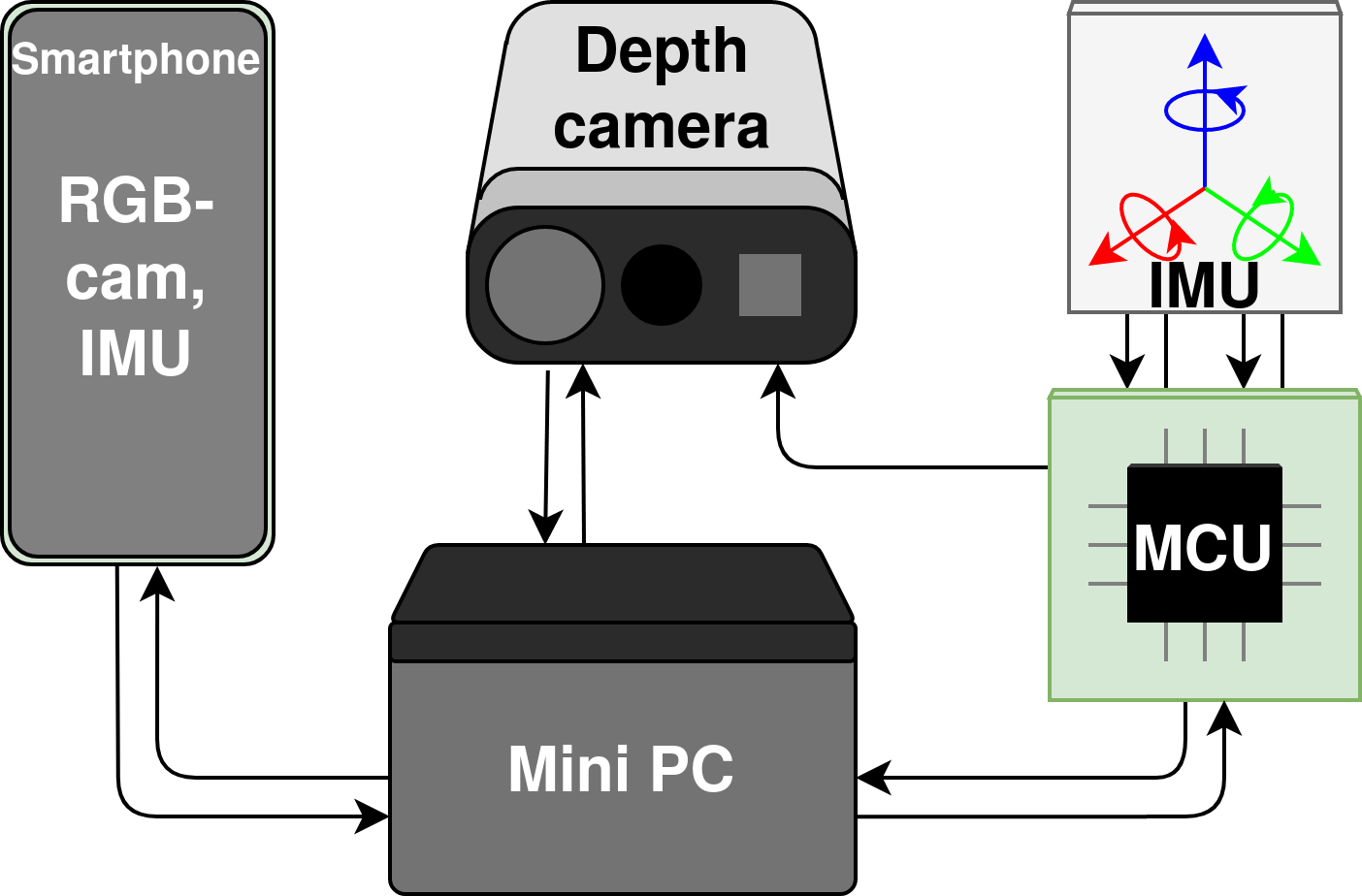}%
    \end{wrapfigure}%
    
    \begin{abstract}
    
    Nowadays, smartphones can produce a synchronized (synced) stream of high-quality data, including RGB images, inertial measurements, and other data.
    Therefore, smartphones are becoming appealing sensor systems in the robotics community. 
    Unfortunately, there is still the need for external supporting sensing hardware, such as a depth camera precisely synced with the smartphone sensors. 
    
    In this paper, we propose a hardware-software recording system that presents a heterogeneous structure and contains a smartphone and an external depth camera for recording visual, depth, and inertial data that are mutually synchronized. 
    The system is synced at the time and the frame levels: every RGB image frame from the smartphone camera is exposed at the same moment of time with a depth camera frame with sub-millisecond precision. We provide a method and a tool for sync performance evaluation that can be applied to any pair of depth and RGB cameras. Our system could be replicated, modified, or extended by employing our open-sourced materials.
    
    
    
    \end{abstract}
    
\end{minipage}}}

\maketitle


\section{Introduction}




A sensor system that consists of various combinations of multiple types of sensors, such as visual or depth cameras, LiDARs, radars, or inertial measurement units (IMUs), generally produces a richer stream of data than a system with a single type of sensor. These data can produce more diverse information that is highly valuable for solving problems of state estimation~\cite{campos2021orb, schops2019bad, shao2019stereo, zhang2015visual} or 3D reconstruction~\cite{dai2017bundlefusion, labbe2019rtab, zollhofer2018state}, among many.
In some of these problems, e.g., Lidar-Inertial Odometry (LIO)~\cite{ye2019tightly, shan2020lio}, the data can be properly fused to obtain more accurate localization and mapping results, while for other problems, part of data can be utilized as reference data for training of machine learning or deep learning algorithms, e.g., in single-image depth estimation~\cite{ignatov2021fast, hu2019revisiting}, or to be used for benchmarking~\cite{dai2017scannet, knapitsch2017tanks, cortes2018advio}.

The quality of sensor synchronization (sync) can severely determine the quality of the obtained results. Poor synchronization directly impacts the ability of the methods to perform sensor fusion: in a moving sensor system, the sync error results in a projection error of the exteroceptive sensors (multiple cameras) due to the parasitic displacement occurred during unwanted time offset between frame shots from different cameras~\cite{olson2010passive}. 
Out of sync can drastically degrade the performance of IMU-based image stabilisation~\cite{english2015triggersync}. Visual-Inertial Odometry (VIO) algorithms cannot correctly track image keypoints when IMU measurements and images are not correctly synced~\cite{tschopp2020versavis}.


\begin{figure*}[t]
    \includegraphics[width=\textwidth]{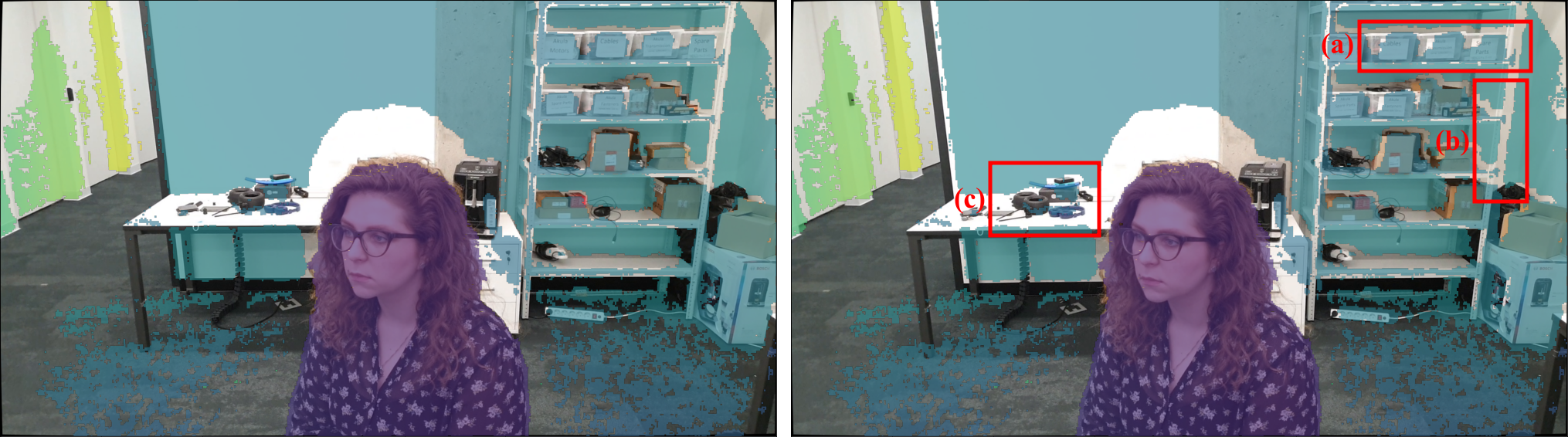}
    \caption{These images show a person's upper body capturing while carrying the system around the person with low dynamic (about 0.2 s$^{-1}$ angular velocity and 1 m/s tangential linear velocity). 
    Every image depicts a depth frame overlaid on a visual frame. The left image frames are synchronized on the frame level, while the frames on the right image are exposed with 33 ms offset. Because some background objects have low reflectance (empty depth frame pixels), it is possible to estimate frame mismatch. Thus labels of the plastic cases on a shelf in the top right corner of the right image (red box (a)) do not coincide with their depth representation; the empty depth frame pixel area related to the right rack of the shelf (b) or a heap of objects on a table (c) has a shift of 5-10 cm from its visual twin. At the same time, there is no such frame mismatch in the left image.}
    \label{fig_sync_example}
\end{figure*}

In recent years, smartphones have become an available option for dedicated sensor systems \cite{grossi2019sensor, callebaut2019bring}; therefore, they began to draw considerable attention in the robotics community thanks to the continuous improvement of their sensor integration. 
A modern smartphone provides tightly synced visual cameras with an IMU sensor making this gadget an ideal independent data gathering and processing unit. However, a smartphone might need some support from external high-quality 
sensors that expand data modality, such as a depth camera to collect 3D-data.

To this end, a smartphone RGB camera must be synced with the depth camera: a delay of only a few milliseconds might result in depth-to-visual frame mismatch of several image pixels that corresponds to reprojection error of several centimeters for objects at the distance of a couple of meters, when capturing a dynamic scene or in case of a moving recording system (Fig.~\ref{fig_sync_example}). 
Unfortunately, smartphones are closed systems that do not allow wired triggering as in most industrial settings. Therefore, combining a smartphone and external sensors requires hybrid hardware-software approaches to sync the overall recording system correctly.

Current solutions of smartphone sync are based on network protocols such as the Precision Time Protocol (PTP)~\cite{ptp2008} or the Network Time Protocol (NTP)~\cite{rfc5905} 
However, for precise sync, the protocols require specific conditions of a network connection, including constraints on latency and symmetry; complete implementation of the sync protocols is not always available in smartphones;
background processes launched on smartphones negatively impact on protocols sync mechanism~\cite{faizullin2021twist}.
Alternatively, data-driven sync methods analyze data properties to align the data from different sensors. However, their direct application provides low (several millisecond) sync precision~\cite{shrestha2006synchronization} or needs auxiliary tools~\cite{bradley2009synchronization} and non-typical events that may not be obtained~\cite{vsmid2017rolling}.

This paper aims to propose a synced recording system that includes both external sensors and sensors embedded in a smartphone.
We call such a system a \textit{heterogeneous system} due to different ways of mutual sync of the sensors and different ways of their integration into the overall system. 
The main content of the SmartDepthSync is aimed at describing the integration and sync of the system. Below, we list the contributions. 

In our first contribution, we propose a multi-modal heterogeneous hardware-software system for data collection that consists of a combination of smartphone RGB-camera and IMU and an individual high-grade depth camera. We made the design of our system publicly available and shared all the materials and code on our project page\footnote{\url{https://github.com/MobileRoboticsSkoltech/bandeja-platform} or \url{https://github.com/Marselka/bandeja-platform}}.
To gather smartphone data, we developed an open-source OpenCamera Sensors mobile application that enforces low-level application programming interface (API) to achieve highly accurate and precise data timing. This application has been made publicly available for the community\footnote{\url{https://github.com/MobileRoboticsSkoltech/OpenCamera-Sensors}}.

In our second contribution, we provide a two-level method for sync of all the sensors: 
(i) \textit{time sync}, where the entire obtained data are presented in a common time domain, 
(ii) \textit{frame sync}~-- smartphone and depth camera frames are exposed at the same moments of time with \textit{sub-millisecond} precision and a low frame sync drift that maintains the precision order for more than 60 minutes. 
The time sync level is obtained by our previously proposed gyroscope based time sync algorithm called Twist-n-Sync~\cite{faizullin2021twist}. Now, we made its implementation publicly available as a Python package\footnote{\url{https://github.com/MobileRoboticsSkoltech/twistnsync-python} or \url{https://github.com/Marselka/twistnsync-python}}.

Our third contribution consists of a novel approach of an RGB to depth camera sync precision evaluation, making use of the depth camera infrared projector features. 

The present 
paper is structured as follows: in Sec.~\ref{sec_rw} we review the related work, Sec.~\ref{sec_overview} overviews the main components of the recording system and Sec.~\ref{sec_soft} describes the software components.
Sec.~\ref{sec_time_sync} describes the time and frame sync process between all the components. Sec.~\ref{sec_eval} evaluates the results and, finally, Sec.~\ref{sec_concl} concludes the paper.
\section{Related Work}\label{sec_rw}
Since the SmartDepthSync lies in the intersection of multiple fields: data acquisition platform, sensors synchronization, smartphone as a sensor, we briefly introduce current research related to our work.

\textbf{Hardware platforms.}
Hardware (HW) sync is the most precise and flexible synchronization method that is frequently utilized for sensor platforms.
In this paragraph, we count state-of-the-art HW platforms for data acquisition.
Tschopp et al.\cite{tschopp2020versavis} propose an open-source sensor suite with HW triggering based on a microcontroller unit (MCU). The system supports a number of sensors, including visual, depth, and inertial sensors. The authors achieve 1 ms sync accuracy and demonstrate sync importance on Visual-Inertial (VI) Simultaneous Localization and Mapping (SLAM), multi-modal mapping, navigation, and reconstruction tasks. Liu et al.\cite{liu2021matter} share principles to follow for sensors sync and consider intra- and inter-machine sync levels. To realize the principles, the authors design a field-programmable gate array (FPGA) based platform for sync of visual cameras, lidar, radar, and IMU. For intra-machine sync they state 100 \si{\micro\second} accuracy (lidar to the rest sensors). Nikolic et~al.~\cite{nikolic2014synchronized} develop an HW-synced VI sensor unit for real-time SLAM based on FPGA. Due to tight integration, inter-sensor time offset achieves down to several \si{\micro\second}. Schuber et al.\cite{schubert2018vidataset} present a VI dataset and provide a description of a VI sensor system (two visual cameras and an IMU) used for data gathering. The system is HW-synced; however, the time offset between the IMU and cameras is estimated by software sync using grid-search. Faizullin et~al.~\cite{faizullin2021opensource} provide an open-source HW system design for IMU-to-lidar sync based on MCU. The system achieves 1 \si{\micro\second} sync precision by emulating GPS clock by MCU in contrast to less precise PTP based sync in~\cite{liu2021matter}. Schneider et~al.~\cite{schneider2010fusing} propose a method for lidar to camera sync that makes a visual camera expose the images when the lidar laser beam coincides with the camera's optical axis.

\textbf{Network-based time sync.}
An interface for HW sensors sync is not always available. The main well-developed alternative approach to sync devices is to use network-based protocols. Those protocols are based on exchanging messages with time among devices in a sensor network and gathering statistics on arrival time between messages. The main representative of this class is the NTP protocol, supported on many operating systems, including embedded devices. Among other protocols and algorithms Simple Network Time Protocol (SNTP)~\cite{rfc4330}, PTP, Reference Broadcast Synchronization (RBS)~\cite{rbs2006}, and Lightweight Time Synchronization (LTS)~\cite{lts2003} provide sub-millisecond clock sync accuracy between devices. The main disadvantage of network-based protocols is the requirements on latency and symmetry of the network connection in order to perform fast and precise sync. Also, an additional barrier of using network protocols is the unavailability of protocol implementation on different platforms and the level of their control from the user side~--- i.e., in the case of Android smartphones, there is no flexible control on sync process from the Android API even though the NTP daemon is provided on the platform.

\textbf{Data-driven time sync.} 
If the techniques above are unavailable or unsuitable, data-driven methods of sync are used as an alternative to them. The data-driven methods utilize properties of data streams from multiple sensors to identify mutual temporal dependence between the streams. For instance, Lukac et al. \cite{lukac2009recovering} applied this concept to the synchronization of seismic sensors during the MesoAmerica Subduction Experiment.
Below, we focus on methods of image frame sync from independent cameras that do not have a common HW trigger or clock. A~number of approaches to synchronize images are based on the analysis of common events captured by the cameras. Abrupt brightness change made by external light sources (natural or artificial) is a perfect companion for sync. Although the need for the light source narrows down the usage of these techniques.
Shrestha~{et al.}~\cite{shrestha2006synchronization} sync video recordings by detecting light intensity change on images produced by still camera flashes. This approach provides about 40 ms accuracy that is dictated by the framing period.
A simple though powerful method for the sync of independent video streams with a precision better than one millisecond is proposed in~\cite{vsmid2017rolling}. 
There, the authors employ the rolling-shutter effect of cameras to increase the time resolution of external flashing events. 
Shrstha~{et al.}~\cite{shrstha2007synchronization} analyze audio tracks to sync independent video records of the same event and achieve sub-frame accuracy.
Bradley~{et al.}~\cite{bradley2009synchronization} set up an external stroboscope to create a common exposure for multiple rolling-shutter cameras. Thus, cameras receive an avalanche of photons reflected from the scene at the same instant of time. This approach is useful in very controlled settings.
In contrast to light-based methods, Caspi~{et al.}~\cite{caspi2006feature} propose a temporal-spatial matching of images from cameras by processing trajectories of moving objects and obtain sub-frame sync accuracy.

\textbf{Smartphones as synchronized sensors.}
Since the quality of modern smartphones is being improved, these devices are becoming more popular as systems of sensors with a wide range of sensing modalities. They can provide more usage when coupled within a network. These types of networks need time sync as well. 
Currently, sync of smartphones is provided by network-based or data-driven techniques.
Callebaut et al.\cite{callebaut2019bring} consider smartphones as sensing platforms and sync them by GPS time data and NTP. Their sync performance analysis shows millisecond order accuracy and precision. Sandha et al.\cite{sandha2019exploiting} carry out an evaluation of smartphone-to-smartphone sync performance based on Bluetooth, Wi-Fi, and audio peripheral and state \mbox{(sub-)millisecond} sync accuracy.
Akiyama et al.\cite{akiyama2015syncsync} apply acoustic-light beacon for smartphone localization utilizing smartphone's camera (rolling shutter effect) and microphone.
Latimer et al.\cite{latimer2014socialsync} --~sub-frame video sync in a smartphone camera network based on NTP.
Ansari et al.\cite{ansari2019wireless} propose an Android app for simultaneous photo shooting from multiple smartphones with sub-millisecond sync accuracy. They use custom NTP in a local net with one smartphone as a server.
In \cite{faizullin2021twist}, the authors propose gyroscope-based smartphones sync technique with microseconds accuracy and implement it for synced photo shooting. The smartphones must be rigidly attached for such high performance.


To the best of our knowledge, none of the existing platforms provide synchronization of heterogeneous sensors, including smartphone and other sensors (RGB, depth camera, lidar, IMU, etc.). In contrast, our platform provides synchronous data gathering, including smartphone video and IMU measurements and standalone depth camera images. Moreover, our technique allows sensors to be synced at sub-frame level~-- it makes RGB and depth frames be exposed at the same instance of time with tens of microseconds precision.
\section{System Overview}\label{sec_overview}

\begin{figure}[t]
    \includegraphics[width=\columnwidth]{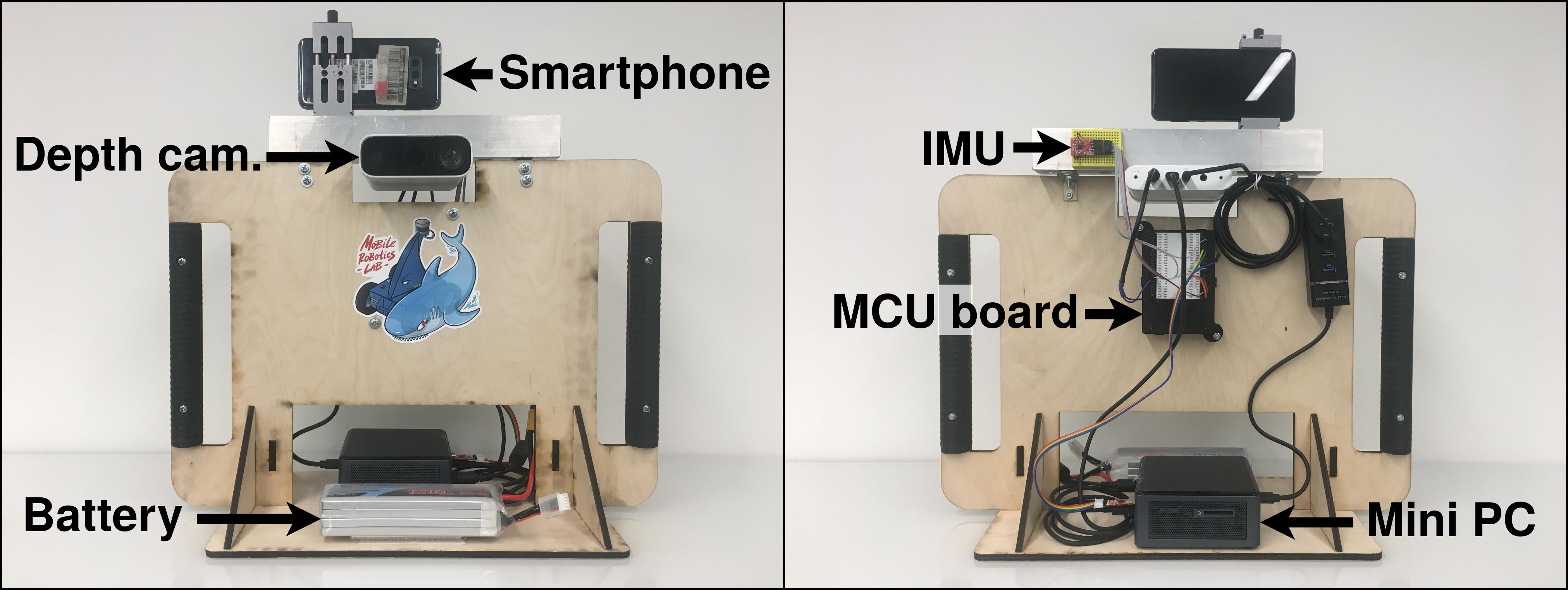}
    \caption{The system common view. All the sensors are attached to the aluminum square tube to have rigid connection.}
    \label{fig_view}
\end{figure}

\begin{figure}[t]
    \includegraphics[width=\columnwidth]{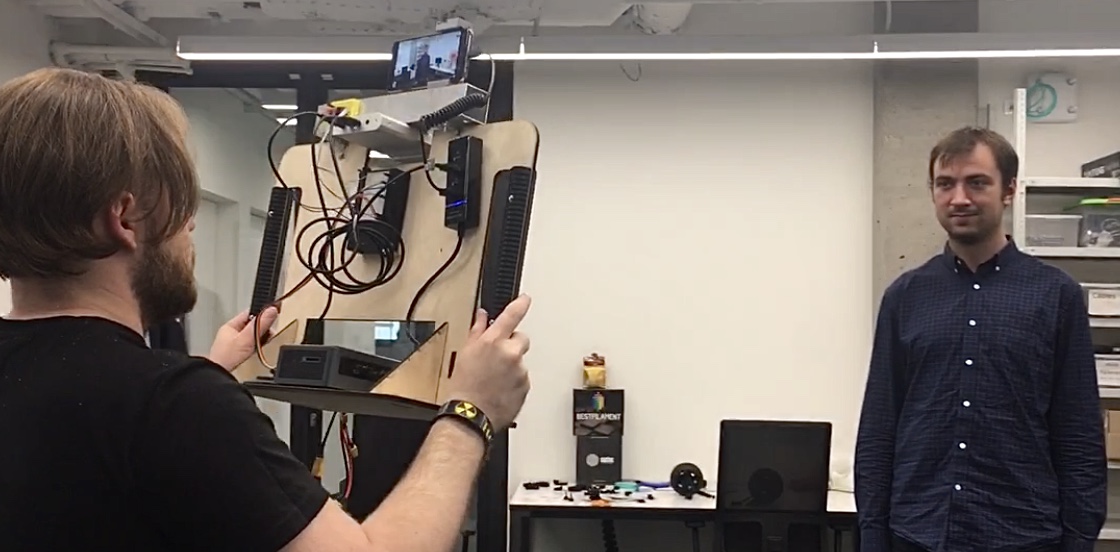}
    \caption{Example of the platform application in a human recording process. The system operator (on the left) moves the system while the laptop operator (not shown) launches and monitors the recording over the SSH connection.}
    \label{fig_record}
\end{figure}

The system is designed to record smartphone videos synchronized at the time and frame levels with depth images from an independent external depth camera. Synchronized IMU measurements are also available. A common view of the system from the front and back sides is presented in Fig.~\ref{fig_view}. The main components of the recording system and their principal roles are as follows:
\begin{itemize}
    \item \textbf{Smartphone} provides data from the \textbf{RGB camera} and \textbf{built-in IMU} (gyroscope, accelerometer).
    \item \textbf{Depth camera} provides high-quality depth data. 
    \item \textbf{Standalone IMU} along with the smartphone built-in IMU provide gyroscope data necessary for the gyroscope-based synchronization algorithm Twist-n-Sync~\cite{faizullin2021twist} between smartphone and depth camera data. Accelerometer data are also available.
    \item \textbf{MCU} acts as a hardware level mediator: it controls triggering of the depth camera frames, obtains standalone IMU measurements, and collects metadata essential for sync.
    \item \textbf{Mini PC} (mini personal computer) controls the recording process, including interaction with the smartphone, depth camera, and MCU.
\end{itemize}

All the system's components are mounted on the hand-held plywood platform designed to carry them comfortably during recordings in the way presented in Fig.~\ref{fig_record}. A metal rig strengthens the platform body to have constant relative transformations between every sensor. 
The system is battery-powered and can work several hours after a full charge.

An additional laptop allows the operator to easily control the recording process via an SSH connection.
During the recording, the smartphone data are saved on the smartphone, whereas other data (depth, inertial data) is saved on the mini PC. Once the recording finishes, all the data are transferred to a common machine, where post-processing re-assigns timestamps into a common time domain.

\begin{figure}[t]
    \centering
    \includegraphics[width=\columnwidth]{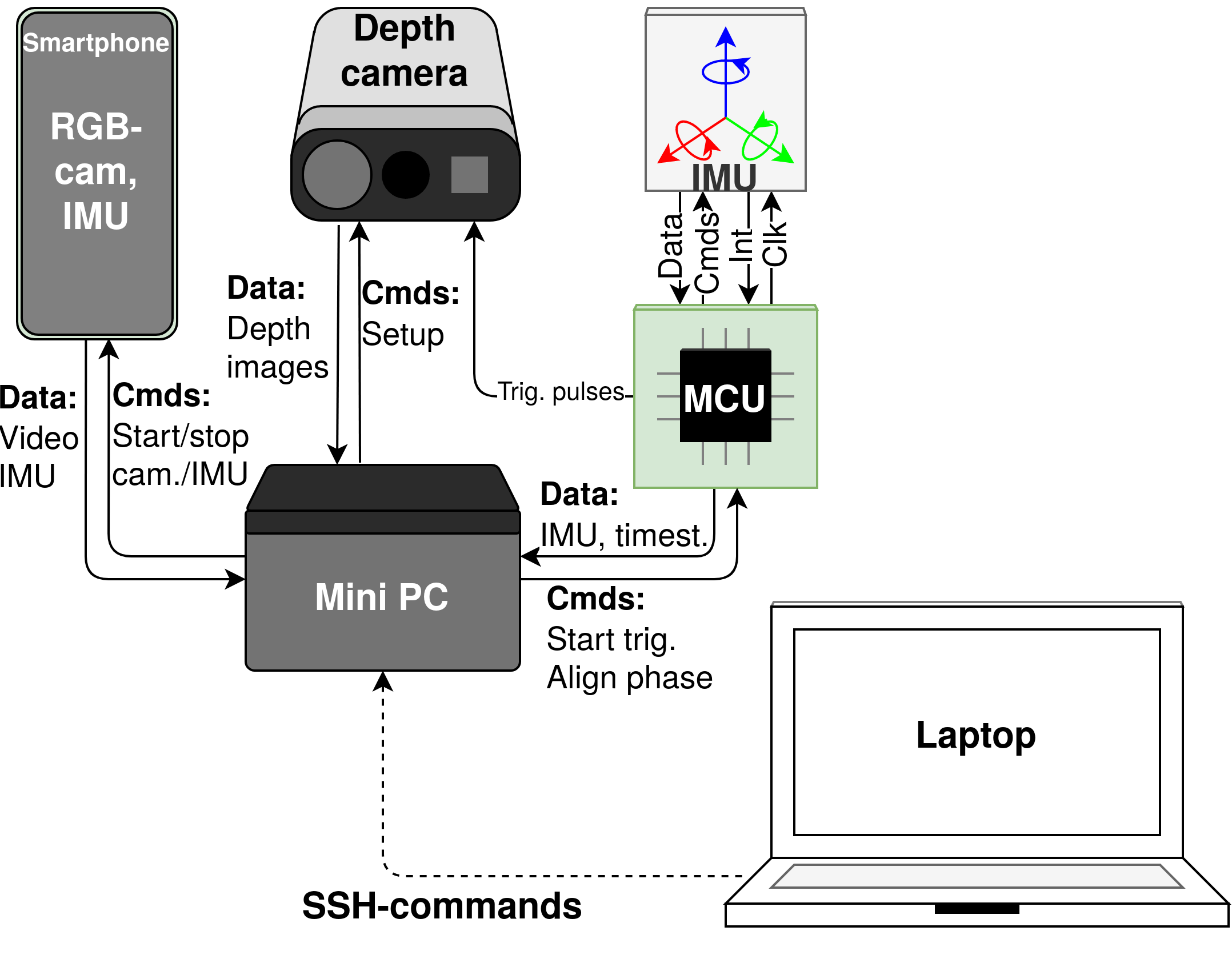}
    \caption{The system topology. Arrows show data/command directions. The depth camera and MCU board are connected to the PC via a USB connection, the smartphone and the laptop are connected to the Wi-Fi network launched on the mini PC.}
    \label{fig_arch}
\end{figure}

\subsection{System Topology}

The block scheme of the system is presented in Fig.~\ref{fig_arch} and shows connections and data flow between the components. The depth camera and standalone IMU (through the MCU board) are connected to the mini PC via a USB interface.
The depth camera is also connected to the MCU board for synced frame triggering. The smartphone is connected to the mini PC via a Wi-Fi network, using an access point launched on the PC. 
The communication between the smartphone and mini PC is provided by our Android OpenCamera Sensors application (Sec.~\ref{subsec_smartapp}) that runs on the smartphone.
Due to remote control API, the app allows a user to run and stop smartphone video and IMU recording over the network commands from mini PC.

\subsection{Hardware Specifics}
\label{sec_hard}

In our system configuration, we use the \textbf{Samsung S10e} smartphone (RGB-camera and IMU), the \textbf{Azure Kinect DK} depth camera (other sensors besides depth are not used), a standalone MPU-9150 IMU-sensor, the Intel(R) NUC mini PC, powered by Ubuntu 18.04 OS, and the STM32F4DISCOVERY MCU board based on STM32F407 MCU. 
The smartphone is powered by Android 10 OS and utilizes a standard  camera for video recording with 1920x1080 pixel resolution and 30 fps frame rate. 
The depth camera is set up to shoot depth images with narrow field of view with 640x576 pixel resolution and a 30 or 5 fps frame rate.
The smartphone and standalone IMU sensors (gyroscope, accelerometer) collect inertial measurements at a 500 Hz data rate.

\section{System Software}\label{sec_soft}

For our system, we have developed the following software components: (i) smartphone application, (ii) MCU firmware, (iii) mini PC recorder, and (iv) data extractor. The first three components are designed for the recording process, the last one is employed in the post-processing step.

\subsection{Smartphone Application}\label{subsec_smartapp}

For recording smartphone data and providing communication between the smartphone and the rest of the system, we developed an open-source OpenCamera Sensors app publicly available on GitHub~\cite{opencamsensors}. It is based on the OpenCamera app~\cite{opencamera} that provides flexible control of camera parameters. OpenCamera Sensors extends it with tools for a synced recording of the camera and IMU data, as well as provides remote control over the Wi-Fi network.

The following features of the OpenCamera Sensors are used in the recording pipeline: (1) start/stop recording video frames and IMU data~--- for data collection, (2) start/stop IMU recording and return recorded data~--- for gyro-based time sync, (3) provide the timestamp of the first video frame at the beginning of the recording~--- for the frame sync.

To improve time and frame sync precision, we adhere to the following setups during every recording:
\begin{enumerate}
    \item We disable sound recording to avoid the addition of undesirable extra frames with unknown timestamps by Android Camera2 API codec.
    \item We lock exposure to achieve a constant frame rate because auto or manual exposure produces a non-stable frame rate.
\end{enumerate}
In addition, we disable optical video stabilization to keep a constant extrinsic relation between the cameras.

\subsection{MCU Firmware}

MCU firmware is developed to perform several tasks aimed at sync and recording: it (i) gathers timestamped inertial data from standalone IMU~--- for MCU to smartphone time sync by Twist-n-Sync, 
(ii) triggers depth camera frames~--- to sync them at the frame level with smartphone RGB-camera frames, 
(iii) collects and sends timestamps of triggering pulses~--- for MCU to depth camera time sync during data extraction.

MCU board provides triggering pulses to a dedicated built-in HW interface of the depth camera with a specific phase for frame sync with the smartphone camera.
This process is discussed in more details in Sec.~\ref{sec_sync_pipeline}.
IMU data gathering process is described in~\cite{faizullin2021opensource}.

\subsection{Mini PC Recorder} \label{sec_mini_pc}
The mini PC recorder software is developed for sync and recording processes. During the sync process, the mini PC interacts with the smartphone and MCU to perform time- and frame-level sync between the smartphone camera and the depth camera, as described in detail in Sec.~\ref{sec_sync_pipeline}. During the recording process, the software spawns multiple ROS nodes that collect depth camera and standalone IMU data and meta-data for post-processing. The collected data are saved to a common rosbag file.

The depth camera data gathering is based on Robot Operating System (ROS) driver~\cite{azureros} that is modified to obtain depth-images with depth camera internal timestamps instead of ROS timestamps. This modification allows sync to avoid intermediate non-precise ROS timestamping~\cite{park2020real, faizullin2021opensource}.
The MCU-IMU ROS driver performs communications with MCU, and it is entirely developed from scratch. All the code is available on our project page.

\subsection{Data Extraction Software}
When data collection is finished, corresponding files are manually uploaded from the smartphone and mini PC to the external machine. Using software for data extraction, they are unpacked, and all the data timestamps are transformed to a single time domain. The tool puts all the extracted files into sub-directories named after corresponding topics. The extractor uses rosbag API to extract messages from the required ROS topics and ffmpeg for smartphone images extraction.  

\section{Synchronization}\label{sec_sync}

The collected data from smartphone and depth cameras has little value if the sensors do not have a common time reference. In order to get the reference, \textbf{time sync} must be performed among the sensors.
However, only time sync is not enough for some tasks where sensors must obtain data at the same moment of time (e.g. RGB-D SLAM~\cite{murORB2}). In particular, for our setup and due to low frame rates of the smartphone and depth camera (30 fps at most), the neighbouring frames from both sensors may be exposed with the offset of up to 16.7~ms (half of the framing period). This mismatch may create adverse artifacts when simultaneously processing visual and depth images that capture dynamic scenes or when the platform is moving (rotating) during the capture.
To overcome this, \textbf{frame sync} also should be performed between the sensors.
In Fig.~\ref{fig_sync_example} an example of the benefit of the frame sync is shown, see the description for details.
This section describes both sync sub-tasks and how they are reflected in the design of our system.

\subsection{Clock Reference}

To perform time sync, a common clock reference should be chosen. Our system has three different clock sources~--- mini PC, smartphone, and MCU. The mini PC clock is not suitable for reference due to its loose relation to the other two clocks: neither the MCU nor the smartphone has a precise sync technique to the mini PC clock. Moreover, the mini PC does not have a precise timestamping procedure for the sensors connected to it (depth camera, standalone IMU). Both the MCU clock and the smartphone clock could be chosen as the reference clock. In our work, we use the MCU-based clock for certainty.

\subsection{Time Sync}
\label{sec_time_sync}

A direct wired time sync of all the system's components is not available because smartphones do not allow for such an interface. 
Common network-based protocols for time sync, such as NTP or PTP, are also not suitable candidates for our system because (1) they rely on network symmetry and constant latency, which neither the PC nor the smartphone guarantee because of uncontrollable background network operations, (2) smartphones cannot provide flexible operation with NTP/PTP daemons. 
Moreover, even in case of solving these two problems and syncing the smartphone and mini PC clocks, network-based protocols cannot solve the depth camera to mini PC relation (the depth camera does not have such a feature), and it makes us return to the ROS timestamping problem discussed in Sec.~\ref{sec_mini_pc}. 
In order to avoid these hindrances, we apply the IMU data-driven solution that syncs smartphone and depth cameras directly, avoiding the mini PC. This solution is described in Sec.~\ref{sec_sync_pipeline}.

\subsection{Frame Sync}
The canonical way to sync data on the frame level is to use hardware triggering of target devices. In our case, the smartphone doesn't provide such an interface, whereas the depth camera supports it. Therefore, to sync devices on the frame level, we use the following approach:
\begin{enumerate}
    \item The smartphone starts video recording and shares a timestamp of the first video frame with the mini PC over the Wi-Fi interface.
    \item Taking into account the timestamp, the mini PC adjusts the phase of the depth sensor using hardware triggering via the MCU.
\end{enumerate}
The phase is defined as the relative offset between the closest smartphone RGB frame exposure middle point and the depth camera frame exposure middle point.
This approach requires both the smartphone and the depth camera to have a constant frame rate. On the smartphone, this property is guaranteed by blocked exposure, on depth camera -- by hardware triggering with a constant period. Moreover, the frame rates must be equal or multiple of each other to keep frames aligned over time.

\subsection{Recording Pipeline with Synchronization}
\label{sec_sync_pipeline}

This sub-section describes the overall recording pipeline and implementation details that provide synced data from the smartphone and the depth camera at the time and frame levels.

After launching the smartphone app, mini PC software, and MCU and depth camera drivers, the system is ready for sync process. 
The whole pipeline from sync up to data extraction is divided into five steps: 
(1) depth camera to MCU time sync,
(2) smartphone to MCU time sync, 
(3) depth to smartphone camera frame alignment, 
(4) recording, 
(5) final timestamps mapping to a common MCU time domain during the extraction.

\textbf{Step 1. Depth camera to MCU time sync.} After the system is launched, the MCU starts to trigger depth camera frames and provides their timestamps in the MCU time domain. We denote the first captured depth frame timestamp in the MCU time domain as $t_{D_0}^M$. 
A unique association of the MCU timestamps with the depth frames that finishes the sync process is done later, during the fifth step.

\textbf{Step 2. Smartphone to MCU time sync.} In order to obtain the time offset between the MCU and smartphone clocks, a data-driven gyroscope-based software sync method is applied. The system is hand-twisted, and both the smartphone and standalone IMUs catch this common event. Then the data from both IMUs with their timestamps are analyzed by the \mbox{Twist-n-Sync} algorithm to obtain the offset denoted as $\Delta t_{SM}$. 
This means that time mapping between the MCU and smartphone is expressed by:
\begin{align}\label{eq:sm_to_mcu}
    t^S = t^M + \Delta t_{SM},
\end{align}
where $t^S$ and $t^M$ is time in the smartphone and MCU time domains respectively.

\textbf{Step 3. Depth to smartphone camera frame alignment.} In this step, smartphone starts video recording and generates the initial frame timestamp in the smartphone time domain $t_{S_0}^S$ that according to (\ref{eq:sm_to_mcu}) is expressed in the MCU time domain by:
\begin{align}
    t_{S_0}^M = t_{S_0}^S - \Delta t_{SM},
\end{align}
where $t_{S_0}^S$ and $t_{S_0}^M$ is the time moment of the initial smartphone frame shot $S_0$ in the smartphone and MCU time domains respectively.

By this moment, the initial smartphone and depth frame timestamps $t_{S_0}^M$ and $t_{D_0}^M$ have been estimated and are now expressed in the same reference frame $M$, the MCU clock domain. We remove the frame superscript for clarity. 
Under the assumption that the smartphone and depth camera framing periods are constant and equal, the exposure moments can be expressed by:
\begin{equation}\label{eq:exposures}
\begin{aligned}
    t_{S_n} &= t_{S_0} + T\,n,\\
    t_{D_m} &= t_{D_0} + T\,m + \Delta t,
\end{aligned}
\end{equation}
where 
$T$ is the framing period,
$n$ and $m$ are frame counts,
$\Delta t$ is the time offset that has to be found for frame alignment, initially it equals zero.
The smartphone and the depth camera are considered aligned if $t_{S_n} = t_{D_m}$ for some pair $(n, m)$; then $\Delta t$ for frame alignment can be found from~(\ref{eq:exposures}) as:
\begin{align}
    \Delta t &= t_{S_0} - t_{D_0} + T\,(n-m).
\end{align}
The $\Delta t$ then to be transformed to the target phase shift $\Delta t_{phase}$ by:
\begin{equation}
\begin{aligned}
    \Delta t_{phase} &= \Delta t \pmod{T}\\
    &= t_{S_0} - t_{D_0}  \pmod{T}.
\end{aligned}
\end{equation}
The obtained $\Delta t_{phase}$ is sent to the MCU that finally corrects the phase on the fly. At this stage, smartphone-to-depth camera frame sync is completed.

MCU HW can set up the phase with a discrete step size equals about 390 ns that is negligible to the overall precision of the sync that is discussed in the next section. 

\textbf{Steps 4 and 5. Recording and extraction.} After obtaining offsets and performing frame sync, the system then records the data.
After recording, during the extraction, data timestamps are finally mapped into the common MCU time domain.
\section{Evaluation}\label{sec_eval}

In this section, we evaluate the precision of the frame alignment, as well as its accuracy drift over time. 





The previous works \cite{ansari2019wireless, vsmid2017rolling, faizullin2021twist} (Sec. \ref{sec_rw}) utilize controlled light sources for evaluation of the sync performance of \textit{rolling shutter} cameras by their methods.
The light sources emit short light flashes (strobes) in specific time moments. The strobes are then caught by the cameras, and the analysis of the synced images is performed. Due to the rolling shutter effect~\cite{vsmid2017rolling}, the strobes are visible only on some pixel rows of images that are exposed at this moment while the rest areas remain darker.
By row position difference between the synced images, one can estimate the time offset between the synced image frames. For this estimation, the row readout time must be known for every image sensor to transform the row position difference to the time offset. More detailed explanation can be found in~\cite{faizullin2021twist}. 
We propose a variation of the above principle in order to measure depth-RGB frame alignment that can be applied to any pair of visual rolling shutter camera and time-of-flight based depth camera. 
In our case, the controlled light source is the depth camera's projector that emits infrared (IR) strobes of light. The strobes are normally used by Azure Kinect DK as a light source in the depth estimation process. The projector generates the strobe train of 9 strobes at every depth image frame exposure, the fifth strobe coincides with the middle point of the exposure time~\cite{azure}.

\begin{figure}[t]
\includegraphics[width=\columnwidth]{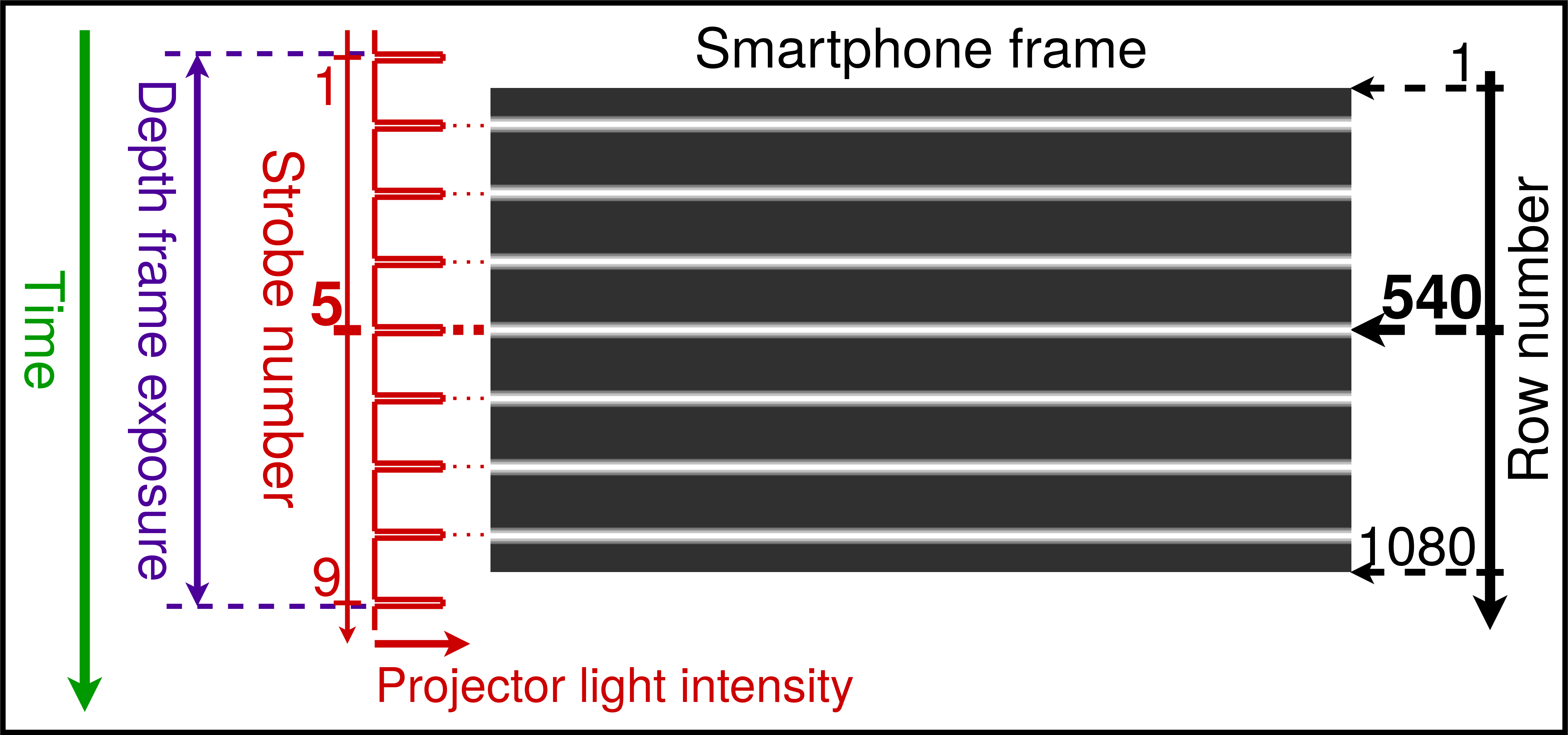}
\caption{Perfect sync case -- the 5th projector strobe coincides with the exposure of the middle (540th) row of the smartphone camera frame. The 1st and 9th strobes are invisible on the frame because the strobe train duration is longer than the entire row-by-row smartphone frame exposure.}
\label{fig_frame_sync}
\vspace{-4.1mm}
\end{figure}

We consider the coincidence of the projector's fifth infrared strobe flash moment with the middle exposure point of the middle row of a smartphone frame as \textit{perfect sync}. The 540th row is chosen as the middle row, additional description can be found in Fig.~\ref{fig_frame_sync}.
Fluctuations of the row position around the middle one determine precision of the smartphone-to-depth cameras frame sync, whereas a steady shift of the mean bright row position over time defines the accuracy drift.

\begin{figure*}[t]
\includegraphics[width=\textwidth]{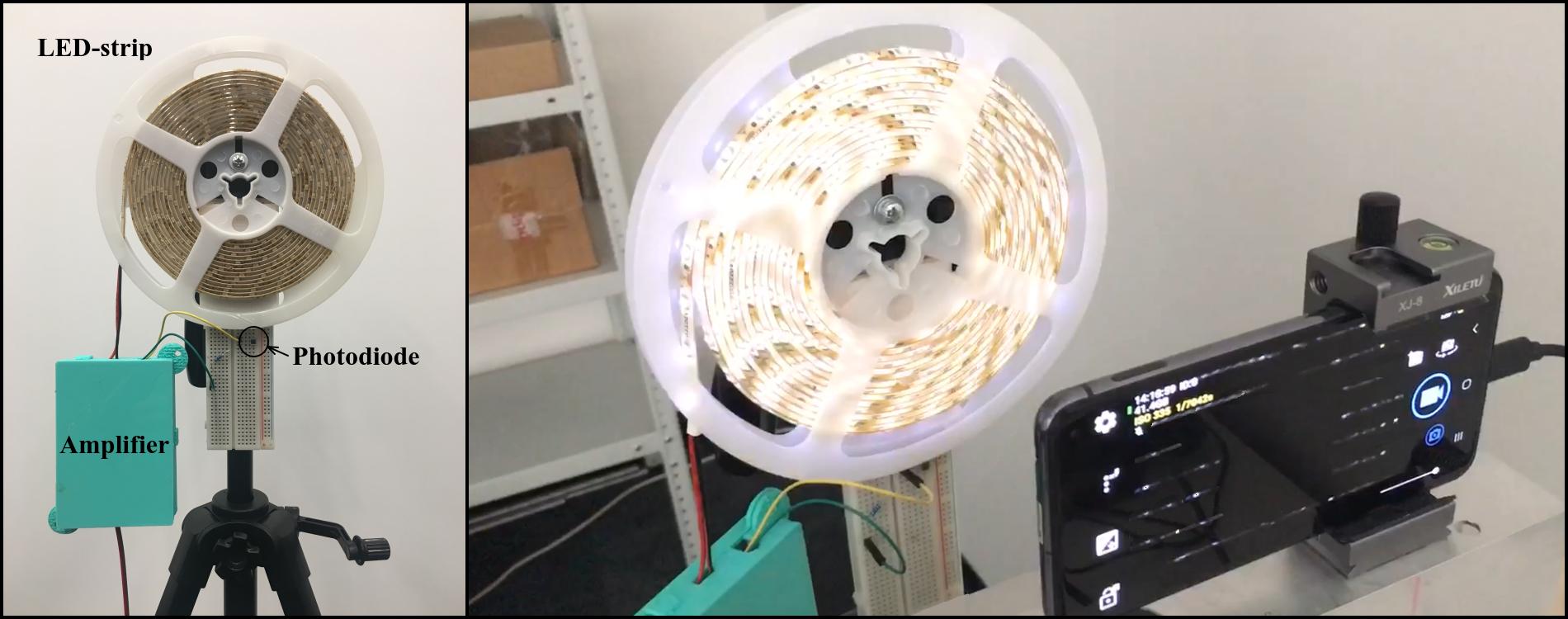}
\caption{Common view of the re-transmitter (left) and experimental setup (right). The amplifier is used to strengthen the photodiode signal to flash the LED-strip. The smartphone camera is set up to the lowest exposure time to resolve the train of short strobes. It could be noticed that no more than seven strobes can be observable on the single smartphone image because the whole image readout time is less than the strobes train duration.}
\label{fig_retrans}
\end{figure*}

The problem of infrared (IR) light is that it is invisible for the smartphone camera image sensor because of the IR filter in the camera's optical scheme.
To bypass this issue, we designed a light re-transmitter that receives the IR strobes and repeats them in the visible spectrum. The re-transmitter transforms IR-light to the electrical signal by a photodiode, then amplifies the signal power to flash a LED-strip. The visible light is then captured by the smartphone camera with no constraints. A common view of the re-transmitter and experimental setup used for the evaluation are shown in~Fig.~\ref{fig_retrans}. 

\begin{figure*}[t]
\includegraphics[width=\textwidth]{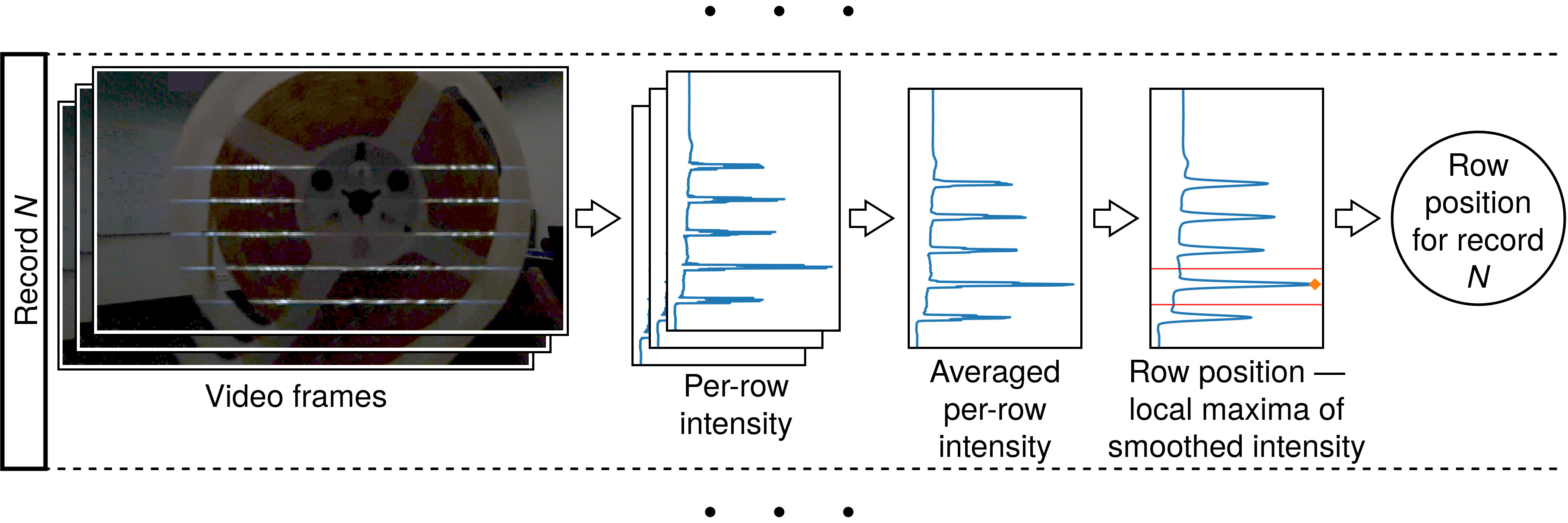}
\caption{Statistics gathering for frame sync precision evaluation. Row position determination is shown for a single record from 52 in total. The number of frames for the determination is 16.}
\label{fig_imagesstat}
\end{figure*}

Below, we firstly evaluate inter-launch sync precision. This means we monitor the row position corresponding to the certain strobe at every recording start for a number of system launches, then we analyze the obtained distribution of inter-launch row positions.
To collect the inter-launch row positions, we recorded 52 short smartphone video sequences and extracted 16 consequent frames from every video, as depicted in Fig.~\ref{fig_imagesstat}.
For each of these frames from a single video, we (i) extract per-row intensity, then (ii) average the intensities over images, and (iii) apply the smoothing to the averaged per-row intensity by the Gaussian filter with a kernel size of 7 lines (or 71.4 {\si{\micro\second}}) to get more robust peak positions.
After smoothing, we choose the region that includes a certain peak for every frame in every recording (the fourth peak is chosen as the highest peak among the others in the experiment) and obtain the peak position for every trial. The obtained peak distribution provides \textit{sub-millisecond} precision, stated in Table~\ref{table_stat}.
The row-to-time transformation is obtained by measuring the average number of rows between every available pair of neighbouring bright peaks (determined by pairs of consecutive strobes in a strobe train), knowing the time delay between the strobes from~\cite{azure}. During the evaluation, we manually injected several milliseconds offset to the depth camera phase to always observe the first strobe on the smartphone images. This is to be sure that the fourth peak is picked up on every image. Otherwise, it is not possible to enumerate strobes because the whole train does not fit into an image (this is described in Fig.~\ref{fig_frame_sync} and can be seen on the smartphone screen in~Fig.~\ref{fig_retrans}).

\begin{figure}[t]
\includegraphics[width=\columnwidth]{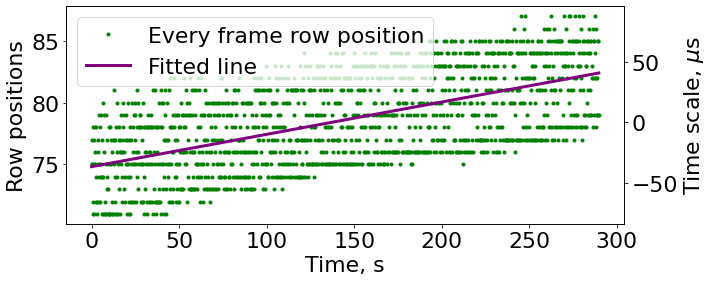}
\caption{Frame sync drift estimation over 5 minutes. For clarity, the right axis indicates time scale of the drift.}
\label{fig_drift}
\vspace{-4.1mm}
\end{figure}

For frame sync accuracy drift evaluation, we recorded a 5-minute video and gathered the pixel row position for some strobe for every sixth smartphone frame. Then linear regression is applied to 2100 pairs (frame timestamp~-- row position) to compute frame sync drift over time. Fig.~\ref{fig_drift} depicts the row position drift. The numerical result is shown in~Table~\ref{table_stat}.
We do not provide the initial sync accuracy value estimation because the offset between smartphone and depth camera exposure starts is set up manually to see the fifth strobe in the middle of the smartphone images. 



%

\begin{table}[h]
    \caption{Sync precision and drift. The interquartile range (IQR) and standard deviation (SD) measure inter-launch sync precision. The accuracy drift over time is for a single recording. Such performance allows recording with sub-millisecond precision during an hour.}
    \label{table_stat}
    \begin{center}
        \begin{tabular}{cc}
            \textbf{IQR(SD)} & \textbf{Accuracy drift}\\
            \hline
            66.3 (61.6) \si{\micro\second} & 16.34 \si{\micro\second}/m
        \end{tabular}
    \end{center}
    \vspace{-4.1mm}
\end{table}

\section{Conclusion}\label{sec_concl}


In this paper, we have proposed an open-source recording system of heterogeneous sensors including, but not limited to smartphone visual camera, standalone depth camera, and IMUs. The main contribution of our work is on the hybrid approach to sync different time authorities: MCU and smartphone in our case, but easily extendable to other configurations. To achieve this, we have combined traditional wired-based trigger sync with software sync for the smartphone. 
This allows us to sync the sensors not only at the time but also at the frame level guaranteeing synchronous frame triggering of the smartphone and depth camera.
The sync performance thanks to the unified sync approach does not degrade when increasing the number of sensors: any number of sensors that support hardware sync can be directly connected to the trigger; any number of sensors that include IMU can be synchronized by our software sync with microseconds accuracy as have been shown in our previous work.

We have shown that our system achieves a sync precision of 66 $\si{\micro\second}$ and can maintain stable sync conditions for recording during an hour due to low sync drift.
In addition, we have proposed a new measurement process to establish precise sync between a smartphone RGB and a standalone depth camera employing properties of a depth camera infrared projector.
We have released a synchronization package and a smartphone recording app; all code, design, and scripts necessary to replicate our recording platform or modify it with new configurations have been made publicly available.




\bibliographystyle{IEEEtran}
\balance
\bibliography{bib/IEEEabrv,bib/bib}



\end{document}